\documentclass{article} 
\usepackage{nips15submit_e,times}
\usepackage{hyperref}
\usepackage{url}
\usepackage{algorithm}
\usepackage{algorithmic}
\usepackage{caption}

\title{Algorithms for Linear Bandits on Polyhedral Sets}

\author{
	Manjesh K Hanawal \\
	Department of ECE\\
	Boston Unversity\\
	Boston, MA 02215 \\
	\texttt{mhanawal@bu.edu} \\
	\And
	Amir Leshem \\
	Department of EE \\
	Bar-Ilan University \\
	Ramat-Gan, Israel 52900\\
	\texttt{leshema@eng.biu.ac.il} \\
	\AND
	Venkatesh Saligrama \\
	Department of ECE\\
	Boston Unversity\\
	Boston, MA 02215 \\
	\texttt{srv@bu.edu} 
}

\nipsfinalcopy 

\usepackage{amsbsy}
\usepackage{amsmath}
\usepackage{amssymb}
\usepackage{graphicx}

\newtheorem{theorem}{Theorem}
\newtheorem{lemma}{Lemma}

\newcommand{\vb}{\mbox{${\bf b}$}}

\newcommand{\vx}{\mbox{${\bf x}$}}

\newcommand{\ve}{\mbox{${\bf e}$}}

\newcommand{\vy}{\mbox{${\bf y}$}}

\newcommand{\mA}{\hbox{{\bf A}}}

\newcommand{\gd}{\delta}

\newcommand{\gth}{\theta}
\newcommand{\gthh}{\mbox{$ \hat \theta$}}

\newcommand{\gn}{\nu}

\def\bm#1{\mbox{\boldmath $#1$}}

\newcommand{\vth}{\mbox{$\bm \theta$}}
\newcommand{\vthh}{\mbox{$\bm {\hat \theta}$}}

\newcommand{\cR}{{\cal R}}

\newtheorem{question}{Question}[section]
\newtheorem{coro}{Corollary}[section]

\newcommand{\beq}{\begin{equation}}
\newcommand{\eeq}{\end{equation}}
\newcommand{\bea}{\begin{array}}
\newcommand{\ena}{\end{array}}
\newcommand{\bds}{\begin {itemize}}
\newcommand{\eds}{\end {itemize}}
\newcommand{\bdf}{\begin{definition}}
\newcommand{\blm}{\begin{lemma}}
\newcommand{\edf}{\end{definition}}
\newcommand{\elm}{\end{lemma}}
\newcommand{\bthm}{\begin{theorem}}
\newcommand{\ethm}{\end{theorem}}
\newcommand{\bprp}{\begin{prop}}
\newcommand{\eprp}{\end{prop}}
\newcommand{\bcl}{\begin{claim}}
\newcommand{\ecl}{\end{claim}}
\newcommand{\bcr}{\begin{coro}}
\newcommand{\ecr}{\end{coro}}
\newcommand{\bquest}{\begin{question}}
\newcommand{\equest}{\end{question}}

\newcommand{\BS}{\boldsymbol}
\newcommand{\OL}{\overline}
\newcommand{\A}{\BS{\alpha}}
\begin{document}

%

\maketitle

\begin{abstract}
We study stochastic linear optimization problem with bandit feedback. The set of arms take values in an $N$-dimensional space and belong to a bounded polyhedron described by finitely many linear inequalities. We provide a lower bound for the expected regret that scales as $\Omega(N\log T)$. We then provide a nearly optimal algorithm that alternates between exploration and exploitation intervals and show that its expected regret scales as $O(N\log^{1+\epsilon}(T))$ for an arbitrary small $\epsilon >0$. We also present an algorithm than achieves the optimal regret when sub-Gaussian parameter of the noise is known. Our key insight is that for a polyhedron the optimal arm is robust to small perturbations in the reward function. Consequently, a greedily selected arm is guaranteed to be optimal when the estimation error falls below some suitable threshold. Our solution resolves a question posed by   \cite{MOR11_LinearlyParametrized_RusmevichientongTsitsiklis} that left open the possibility of efficient algorithms with asymptotic logarithmic regret bounds. We also show that the regret upper bounds hold with probability $1$.  Our numerical investigations show that while theoretical results are asymptotic the performance of our algorithms compares favorably to state-of-the-art algorithms in finite time as well. 
\end{abstract}

\section{Introduction}
\label{sec:Intro}
Stochastic bandits are sequential decision making problems where a learner plays an action in each round and observes the corresponding reward. The goal of the learner is to collect as much reward as possible or, alternatively minimize regret over a period of $T$ rounds.
{\em Stochastic linear bandits} are a class of {\em structured bandit problems} where the rewards from different actions are correlated. In particular, the expected reward of each action or arm is expressed as an inner product of a feature vector associated with the action and an unknown parameter which is identical  for all the arms. With this structure, one can infer reward of arms that are not yet played from the observed rewards of other arms. This allows for considering cases where number of arms can be unbounded and playing each arm is infeasible.

Stochastic linear bandits have found rich applications in many fields including web advertisements \cite{WWW10_Contextaulbandits_LiChuWei}, recommendation systems \cite{Book_RecommenderSystem_ZankerFelferningFriedtich}, packet routing, revenue management, etc.
In many applications the set of actions are often defined by a finite set of constraints. For example, in packet routing, the amount of traffic to be routed on a link is constrained by its capacity. In web-advertisements problems, the budget constraints determine the set of available advertisements. It follows that the each arm in these applications belongs to a polyhedron.

Bandit algorithms are evaluated by comparing their cumulative reward against the optimal achievable cumulative reward and the difference is referred to as regret. The focus of this paper is on characterizing asymptotic bounds for regret for fixed but unknown reward distributions, which are commonly referred to as problem dependent bounds~\cite{COLT08_StochasticLinearOptimization_DaniHayesKakad}. 

We consider linear bandits where the arms take values in an $N$-dimensional space and belong to a bounded polyhedron described by finitely many linear inequalities. We derive an asymptotic lower bound of $\Omega(N\log T)$ for this problem and present an algorithm that is (almost) asymptotically optimal. Our solution resolves a question posed by   \cite{MOR11_LinearlyParametrized_RusmevichientongTsitsiklis} that left open the possibility of efficient algorithms with asymptotic logarithmic regret bounds. Our algorithm alternates between exploration and exploitation phases, where a set of arms on the boundary of the polyhedron is played in exploration phases and a greedily  selected arm is played super-exponentially many times in the exploitation phase.  Due to the simple nature of the strategy we are able to provide upper bounds which hold almost surely. We show that our regret concentrates around its expected value with probability one for all $T$. In contrast regret for upper confidence bound based algorithms concentrates only at a polynomial rate \cite{TCS09_ExplorationExploitationTradeOff_AudibertMunosSzepes}. Thus, our algorithms are more suitable for risk-averse decision making. A summary of our results and comparison of regrets bounds is given in Table \ref{tab:RegretComparison}. Numerical experiments show that its regret performance compares well against state-of-the-art linear bandit algorithms even for reasonably small rounds while being significantly better asymptotically.
\vspace{-.3cm}
\begin{center}
	\begin{tabular}{ c | c| c || c |c }
		\hline
		& \multicolumn{2}{l}{ $K$-armed bandits}  & \multicolumn{2}{c}{Linear bandits} \\  \hline \hline
		&  dependent  &  independent &  dependent &   independent  \\ \hline
		Lower bounds & $K\log T$ &  $\sqrt{KT}$ &  { $\boldsymbol{N\log T}$} &  $N\sqrt{T}$  \\ \hline
		Upper bounds & $K\log T$ & $\sqrt{KT}$ &  $\boldsymbol{N\log^{1+\epsilon} T}$ &  $N\sqrt{T}$   \\ \hline
		Efficient  algorithm & UCB1 \cite{ML02_FiniteTimeAnalysis_AuerBianchiFischer}  & MOSS \cite{JMLR2010_RegretBoundsAndMinimax_AudibertBubeck} & \textbf{SEE} (this paper) & $ConfidenceBall_2$ \cite{COLT08_StochasticLinearOptimization_DaniHayesKakad}  \\ \hline
		\label{tab:RegretComparison}
	\end{tabular}
	\vspace{-.5cm}
	\captionof{table}{Summary of (problem) dependent and (problem) independent regret bounds in multi-armed bandits and linear bandits. We considered linear bandits over a bounded subset of N-dimensional subspace with $\Delta>0$. The column with bold letters presents the bounds obtained in this paper.}
\end{center}

\noindent
{\bf Related Work:}
Our regret bounds are related to those described in \cite{COLT08_StochasticLinearOptimization_DaniHayesKakad}, who describe an algorithm ($ConfidenceBall_2$) with regret bounds that scale as $O((N^2/\Delta) \log^3 T)$, where $\Delta$ is the reward gap defined over extremal points. These algorithms belong to the class of so called OFU algorithms (optimism in the face of uncertainty). Since OFU algorithms play only extremal points (arms), one may think that $\log T$ regret bounds can be attained for linear bandits by treating them as $K$-armed bandits, were $K$ denotes the number of extremal points of the set of actions. This possibility arises from the classical results on the $K$-armed bandit problem due to Lai and Robbins \cite{APM85_Asymptotically_LaiRobbins} who provided a complete characterization of expected regret by establishing a problem dependent lower bound of $\Omega(K\log T)$ and then providing an asymptotically (optimal) algorithm with a matching upper bound.
But, as noted in \cite{MOR11_LinearlyParametrized_RusmevichientongTsitsiklis}[Sec 4.1, Example 4.5], the number of extremal points can be exponential in $N$, and this renders such adaptation of multi-armed bandits algorithm inefficient. In the same paper, the authors pose it as an open problem to develop efficient algorithms for linear bandits over polyhedral set of arms that have logarithmic regret. They also remark that since convex hull of a polyhedron is not strongly convex, regret guarantees of their PEGE (Phased Exploration Greedy Exploitation) algorithm does not hold.

Our work is close to FEL (Forced Exploration for Linear bandits) algorithm developed in \cite{COLT09_ForcedExplorationBased_YadkoriAntosSzepe}. FEL separates the exploration and exploitation phases by comparing the current round number against a predetermined sequence. FEL plays randomly selected arms in the exploration intervals and greedily selected arms in the exploitation intervals. However, our policy differs from FEL as follows-- 1) we always play fixed set of arms (deterministic) in the exploration phases. 2) noise is assumed to be bounded in \cite{COLT09_ForcedExplorationBased_YadkoriAntosSzepe}, whereas we consider more general sub-Gaussian noise model 3) unlike FEL, our policy does not require computationally costly matrix inversions. FEL provides expected regret guarantee of only  $\mathcal{O}(c\log^2T)$ whereas our policy PolyLin has optimal $\mathcal{O}(N\log T)$ regret guarantee. Moreover, the authors in \cite{COLT09_ForcedExplorationBased_YadkoriAntosSzepe} remark that the leading constant $c$ in their regret bound can be set proportional to $\sqrt{N}$ (see discussion following Th 2.4 in \cite{COLT09_ForcedExplorationBased_YadkoriAntosSzepe}), but this seems incorrect in light of the lower bound of $\Omega(N\log T)$ we establish in this paper.

In contrast to the asymptotic setting considered here, much of the machine learning literature deals with problem independent bounds. These  bounds on regret apply in finite time and for the minimax case, namely, for the worst-case over all reward (probability) distributions. \cite{Book_IPredictionLearningAndGames_BianchiLugosi} established a problem independent lower bound of $\Omega(\sqrt{KT})$ for multi-armed bandits, and was shown to be achievable in \cite{JMLR2010_RegretBoundsAndMinimax_AudibertBubeck}. For linear bandits, problem dependent bounds and well studied and stated in terms of dimension of the set of arms rather than its size. In \cite{SIAMR02_ANonStochasticMultiArmed_AuerBiachiFreud}, for the case of finite number of arms, a lower bound of $\Omega(\sqrt{NT})$ with matching upperbounds is established, where $N$ denotes the dimension of the set of arms. 
For the case when the number of arms is infinite or form a bounded subset of a $N$-dimensional space, a lower bound of $\Omega(N\sqrt{T})$ is established in \cite{COLT08_StochasticLinearOptimization_DaniHayesKakad,MOR11_LinearlyParametrized_RusmevichientongTsitsiklis} with matching achievable bounds. 

Several variants and special cases of stochastic linear bandits are available depending on what forms the set of arms. The classical stochastic multi-armed bandits introduced by Robbins \cite{AMS1952_SomeAspectsOfSequenntial_Robbins} and later studied by Lai and Robbins \cite{APM85_Asymptotically_LaiRobbins} is a special case of linear bandits where the set of actions available in each round is the standard orthonormal basis. Auer \cite{JMLR02_UsingConfidenceBounds_Auer} first studied stochastic linear bandits as an extension of ``associated reinforcement learning'' introduced in \cite{ICML99_AssociativeReinformentLearning_AbeLong}. Since then several variants of the problems have been studied motivated by various applications. In \cite{WWW10_Contextaulbandits_LiChuWei,AISTATS2011_ContextualBandits_ChuLiReyzinSchapire}, the linear bandit setting is adopted to study content-based recommendation systems where the set of actions can change at each round  (contextual), but their number is fixed. Another variant of linear bandits with finite action set are {\em spectral bandits} \cite{ICML14_SpectralBandits_ValkoMunos,ICML15_CheapBandits_HanawalSaligramaValko}, where the graph structure defines the set of actions and its size. Several authors \cite{COLT08_StochasticLinearOptimization_DaniHayesKakad,MOR11_LinearlyParametrized_RusmevichientongTsitsiklis,COLT09_ForcedExplorationBased_YadkoriAntosSzepe} have considered linear bandits with arms constituting a (bounded) subset of a finite-dimensional vector space and remains fixed over the learning period.  \cite{NIPS2011_ImprovedAlgorithms_AbbasiPalSzepes} considers cases where the set of arms can change between the rounds but must belong to a bounded subset of a fixed finite-dimensional vector space. 

The paper is organized as follows: In Section \ref{sec:ProbFormulation},  we describe the problem and setup notations. In Section \ref{sec:MainResult}, we derive a lower bound on expected regret and describe our main algorithm SEE and its variant SEE2. In Section \ref{sec:Analysis}, we analyze the performance of SEE, and its adaptation for general polyhedron is discussed in Section\ref{sec:GeneralPoly}. In Section \ref{sec:HighConfidence} we provide probability $1$ bounds on the regret of SEE. Finally, we numerically compare performance of our algorithm against sate-of-the-art in \ref{sec:Experiments}.
\section{Problem formulation}
\label{sec:ProbFormulation}
We consider a stochastic linear optimization problem with bandit feedback over a set of arms defined by a polyhedron. Let  $\mathcal{C} \subset \cR^N$ denote a bounded polyhedral set of arms given by
\beq
\mathcal{C}=\left\{\vx \in \cR^N: \mA \vx \le \vb  \right\}
\eeq
where $\mA \in \cR^{M \times N}, \vb \in \cR^M$. At each round $t$, selecting an arm $x_t \in \mathcal{C}$ results in reward $r_t(\vx_t)$. We investigate the case where the expected reward for each arm is a linear function regardless of the history. I.e., for any history $\mathcal{H}_t$, there is a parameter $\vth \in [-1,1]^N$, fixed but unknown, such that
\[\mathbb{E}[r_t(\vx) | \mathcal{H}_t ]=\vth^\prime\vx  \quad \text{for all \;} t \text{\;and\;} \vx \in \mathcal{C}. \]
Under these setting the noise sequence $\{\gn_t\}_{t=1}^{\infty}$, where
$\gn_t= r_{t}(\vx) -\vx^\prime\vth$
forms a martingale difference sequence. Let $\mathcal{F}_t=\sigma\{\gn_1, \gn_2, \cdots,\gn_t, \vx_1, \cdots,\vx_{t+1}\}$ denote the $\sigma$-algebra generated by noise events and arms selections till time $t$. Then $\gn_t$ is $\mathcal{F}_{t}$-measurable and we assume that it satisfies
\begin{equation}
 \label{eqn:Subgaussian}
\text{for all\;} b \in \mathcal{R}^1 \quad \mathbb{E}[e^{b \gn_t} | \mathcal{F}_{t-1} ]\leq \exp\{b^2R^2/2\},
\end{equation}
i.e., noise  is conditionally $R$- sub-Gaussian which automatically implies $\mathbb{E}[\gn_t | \mathcal{F}_t]=0$  and $\mathbf{Var}(\gn_t) \leq R^2$.  We can think of $R^2$ as the conditional variance of noise. An example of $R$-sub-Gaussian noise is $\mathcal{N}(0,R^2)$, or any bounded distribution over an interval of length $2R$ and zero mean. In our work, $R$ is fixed but unknown.

A policy $\phi:=(\phi_1, \phi_2,\cdots)$ is a sequence of functions $\phi_t: \mathcal{H}_{t-1} \rightarrow \mathcal{C}$ such that an arm is selected in round $t$ based on the history $\mathcal{H}_{t-1}$. Define expected (pseudo) regret  of policy $\phi$ over $T$-rounds as:
\beq
\label{eqn:pseudo}
R_T\left(\phi \right)=T\vth^\prime \vx^* -E\left [ \sum_{t=1}^T\vth^\prime\phi(t) \right]
\eeq
where $\vx^*=\arg\max_{\vx \in \mathcal{C}} \vth^\prime\vx$ denotes the optimal arm in $\mathcal{C}$, which exists and is an extremal point\footnote{Extremal point of a set is a point that is not a proper convex combination of points in the set.} of the polyhedron $\mathcal{C}$ \cite{Book_IntroductionLinear_BertsimasTsitsiklis}. The expectation is over the random realization of the arm selections induced by the noise process.  The goal is to learn a policy that keeps the regret as small as possible.  We will be also interested in regret of the policy defined as
\beq
\label{eqn:regret}
\overline{R}_T\left(\phi \right)=T\vth^\prime \vx^* - \sum_{t=1}^T\vth^\prime\phi(t).
\eeq
For the above setting, we can use $ConfidenceBall_2$  \cite{COLT08_StochasticLinearOptimization_DaniHayesKakad} or $UncertainityEllipsoid$ \cite{MOR11_LinearlyParametrized_RusmevichientongTsitsiklis}  and achieve optimal regret of order $N\sqrt{T}$. For linear bandits over a set with finite number of extremal points, one can also achieve regret that scales more gracefully, growing logarithmically in time $T$, using algorithms for the standard multi-armed bandits. Indeed, from fundamentals of linear programming \[ \arg\max_{\vx \in \mathcal{C}} \vth^\prime \vx= \arg \max_{\vx \in \mathcal{E}(\mathcal{C})} \vth^\prime\vx
, \] where $\mathcal{E}:=\mathcal{E}(\mathcal{C})$ denotes the set of extremal points of $\mathcal{C}$. Since the set of extremal points is finite for a polyhedron, we can use the standard Lai and Robbin's algorithm \cite{APM85_Asymptotically_LaiRobbins}  or UCB1 in \cite{ML02_FiniteTimeAnalysis_AuerBianchiFischer} treating each extremal point as an arm and obtain regret bound (problem dependent) of order $\frac{|\mathcal{E}|}{\Delta}\log T$, where $\Delta:= \vth^\prime \vx^* - \max_{\mathcal{E}\backslash\vx^* }\vth^\prime\vx$ denotes the gap between the best and the next best extremal point. However, the leading term in these bounds can be exponential in $N$, rendering these algorithm ineffective. For example, the number of extremal points of $\mathcal{C}$  can be of the order $ \binom{M+N}{M} =\mathcal{O} ((2N)^M)$. Nevertheless, in analogy with the problem independent regret bounds in linear bandits, one wishes to derive problem dependent logarithmic regret where the dependence on set of arms is only linear in its dimension. Hence we seek an algorithm with regret of order $N\log T$.

In the following,  we first derive a lower bound on the expected regret and develop an algorithm that is (almost) asymptotically optimal.
\section{Main results}
\label{sec:MainResult}
In this section we provide a lower bound on the expected regret and present our proposed policy and prove the main results regarding its complexity.

\subsection{Lower Bound}
\label{subsec:LowerBound}
We establish through a simple example that regret of any asymptotically optimal linear bandit algorithm is lower bounded as $\Omega(N\log T)$. Recall the fundamental property of the linear optimization that an optimal point is always an extremal point. Then any linear bandit algorithm on a polyhedral set of arms always play the extremal points. We exploit this fact, and mapping the problem to a standard multi-armed bandits we obtain the lower bound.

We need the following notations to prove the result. Let $\{\eta(\beta)\}_{\beta \in [0,1]}$ denote a set of distributions parametrized by $\beta \in [0,1]$ and such that each $\eta(\beta)$ is absolutely continuous with respect to a positive measure $m$ on $\mathcal{R}$. Let $p(x;\beta)$ denote the probability density function associated with distribution $\eta(\beta)$, and let $KL(\beta_1,\beta_2)$ denote the Kullback-Leibler (KL) divergence  between distributions $\eta(\beta_1)$ and $\eta(\beta_2)$ defined as $KL(\beta_1,\beta_2)= \int_x p(x;\beta_1)\log \frac{p(x;\beta_1)}{p(x;\beta_2)}m(\rm{d}x)$. Consider a set of $K$ arms. We say that arm $k$ is parametrized by $\beta_k$ if its reward is distributed according to $\eta(\beta_k)$.

We are now ready to state asymptotic lower bound for the linear bandit problem over any bounded polyhedron with positive measure . Without loss of generality, we restrict our attention to uniformly good policies as defined in \cite{APM85_Asymptotically_LaiRobbins}. We say that a policy $\phi$ is uniformly optimal if for all $\vth \in \Theta $, $R(T,\phi)=o(T^\alpha)$ for all $\alpha > 0$.

\begin{theorem}
Let  $\phi$ any uniformly good policy on a bounded polyhedron with positive measure.  For any $\vth \in [0,1]^N$, let $\mathbb{E}[\eta(\theta_k)]=\theta_k$ for all $k=1,2,\cdots, N$. Then,
\begin{equation}
\label{thm:LowerBound}
\liminf_{T \rightarrow \infty} \frac{R_T(\phi)}{\log T} \geq \frac{(N-1) \Delta}{\displaystyle \max_{k: \theta_k < \theta^*}KL(\theta^*,\theta_k)} \quad \mbox{where\;} \quad \theta^* =\arg\max_n \theta_n
\end{equation}
\end{theorem}
{Proof sketch:} First, note that number of extremal points of any bounded polyhedron with positive measure is atleast  $(N+1)$.  We can then restrict to a bounded polyhedron with $N+1$ extremal points. Let $\tilde{\mathcal{C}}=\{\vx \in \mathcal{R}^N: 0 \leq x_i \leq 1\; \forall \;i=1,2\cdots,N\}$. The $(N+1)$ extremal points of $\tilde{\mathcal{C}}$ are $\{\ve_n:n=1,2,\cdots,N\}\cup{\boldsymbol\{0\}}$.  In the linear bandit problem with unknown parameter $\vth$, playing the extremal point $\ve_n$ gives mean reward $\theta_n$. Also, by the property of linear optimization, any OFU policy will only play extremal points in every round.  Then, the linear bandit over polyhedron $\tilde{\mathcal{C}}$ is the same as $N+1$-armed bandit where reward of $k$th arm $k=1,2\cdots,N$ is distributed as $\eta(\theta_k)$ with mean $\theta_k$, and the reward of $N+1$th arm is distributed as $\eta(0)$ with  mean $0$.

The result follows from Lai-Robbin's lower bound for stochastic multi-armed bandits proved in \cite{APM85_Asymptotically_LaiRobbins} after verifying that the mean values of the parametrized distribution satisfy the required conditions.

\subsection{Algorithms}
\label{subsec:Algortihm}
The basic idea underlying our proposed technique is based on the following observations for linear optimization over a polyhedron. 1) The set of extremal points of polyhedron is finite and hence $\Delta >0$. 2) When $\vthh$ is sufficiently close to $\vth$, then over the set $\mathcal{C}$ both $\arg\max \vth^\prime \vx$ and $\arg\max \vthh^\prime \vx$ give the same value.
We exploit these observations and propose a two stage technique, where we first estimate $\vth$ based on a block of samples and then exploit it for much longer block. This is repeated with increasing block lengths so that at each point the regret is logarithmic.
For ease of exposition, we first consider the polyhedron that contains origin and postpone the  general case to Section \ref{sec:GeneralPoly}. 

Assume that the polyhedron $\mathcal{C}=\left\{\vx \in \cR^N: \mA \vx \le \vb  \right\}$ contains origin as an interior point. Let $\ve_n$ denote $n$th standard unit vector of dimension $N$. For all $1\leq n\leq N$, let $\overline{z}_n=\max \left\{z\ge 0,  z\ve_n \in \mathcal{C} \right\}$. The subset of arms $B:=\{\overline{z}_n \ve_n:n=1,2\cdots,N\}$ are the vertices of the largest simplex bounded in $\mathcal{C}$. Since $\gth_n=\vth^\prime \ve_n$ we can estimate $\gth_n$ by repeatedly playing the arm $\overline{z}_n \ve_n$. One can also estimate $\theta_n$ by playing an interior point $z\ve_n \in \mathcal{C}$ for some $z >0$. But as will see later selecting the maximum possible $z$  improves the probability of estimation error.


\begin{minipage}{7.6cm}
{\bf Algorithm-SEE}\\
In our policy- which we refer as  \textbf{S}equential-\textbf{E}stimation-\textbf{E}xploitation (SEE)- we split the time horizon into cycles and each cycle consists of an exploration interval followed by an exploitation interval. We index the cycles by $c$ and denote the exploration and exploitation intervals in cycle $c$ as $E_c$ and $R_c$, respectively. In the exploration interval $E_c$, we play each arm in $\mathcal{B}$ repeatedly for $(2c+1)$ times. At the end of $E_c$, using the rewards observed for each arm in $\mathcal{B}$ in the past $c$- cycles we compute ordinary least square (OLS) to estimate each component $\theta_n, n=1,2,\cdots,N$ separately and obtain the estimate $\hat{\vth}(c)$. Using $\hat{\vth}(c)$ as a proxy for $\vth$, we compute a greedy arm $\vx(c)$ by solving a linear program and play it repeatedly for $2^{c^2/(1+\epsilon)}$ times in the exploitation interval $R_c$, where $\epsilon>0$ in an input parameter. We repeat the process for each cycle. A formal description of SEE is given in the adjacent figure. The estimation in line $13$ is computed for all $n=1,2,\cdots, N$ as follows:
	\beq
	\label{eqn:EstimateTheta}
	\gthh_n(c)=\frac{1} {c^2} \sum_{i=0}^{c} \sum_{j=1}^{2i+1} r_{t_{i,n,j}}/z_{n},
	\eeq
\end{minipage}\hspace{.3 cm}
\begin{minipage}{5.8cm}
	\begin{algorithm}[H]
		\caption{SEE}
		\label{agm:PEGE-PLB2}
		\begin{algorithmic}[1]
			\STATE \textbf{Input:}
			\STATE $\mathcal{C}$: The polyhedron
			\STATE $\epsilon$: Algorithm parameter
			\STATE \textbf{Initialization:}
			\STATE  Compute the set $\mathcal{B}$
			\FOR {$c = 0,1,2,\cdots$}
			\STATE \textbf{Exploration:}
			\FOR {$n=1\to N$}
			\FOR {$j=1 \to 2c+1$}
			\STATE Play arm $z_n\ve_n \in \mathcal{B}$, \\ observe reward $r_{t_{c,n,j}}$
			\ENDFOR
			\STATE Compute $\hat{\theta}_n(c)$
			\ENDFOR 		
			\STATE $\hat{\vth}(c)\leftarrow (\hat{\theta}_1(c),\hat{\theta}_2(c)\cdots, \hat{\theta}_N(c))$
			\STATE $\vx(c)\leftarrow \displaystyle \arg\max_{\vx \in \mathcal{C}} \vx^\prime \hat{\vth}(c) $
			\STATE \textbf{Exploitation:}
			\FOR {$j=1 \to \lfloor 2^{c^2/1+\epsilon} \rfloor$}
			\STATE Play arm \vx(c), observe reward
			\ENDFOR
			\ENDFOR
		\end{algorithmic}
	\end{algorithm}
\end{minipage}

Note that in the exploration intervals, SEE plays a fixed set of arms and no adaption happens, adding positive regret in each cycle. The regret incurred in the exploitation intervals starts reducing as the estimation error gets small, and when it falls below $\Delta/2$ the step (line-16) selects the optimal arm and no regret is incurred in the exploitation intervals (Lemma \ref{lma:InfinityErrorBound}). As we will show later, the probability of estimation error decays super-exponentially across the cycles, and hence the probability of playing a sub-optimal arm in the exploitation interval also decays super-exponentially.

\begin{theorem}
	\label{thm:ALGO1}
	Let the noise be $R$-sub-Gaussian and without loss of generality\footnote{For general $\vth$, we replace it by $\frac{\vth}{ \| \vth \|_\infty}$ and the same method works. Only $R_m$ is scaled by a constant factor.} assume $\vth \in [-1, 1]^N$. Then, the expected regret of SEE, with parameter $\epsilon>0$ is bounded as follows:
	\begin{equation}
	\label{eqn:RegAlg2}
     R_T(SEE) \leq 2R_m N\log^{1+\epsilon}T+ 4R_mN\gamma_1,
	\end{equation}
	where $R_m$ denotes the maximum reward.  $\gamma_1$ is a constant that depends on noise parameter $R$ and the sub-optimality gap $\Delta$.
\end{theorem}

The $\epsilon$ parameter determines the length of the exploitation intervals, and larger $\epsilon$ implies that SEE spends less time in exploitation and more time in exploration. Increasing $\epsilon$ will make SEE spend more time in explorations resulting in improved estimations and reduces the probability of playing sub-optimal arm in the exploitation intervals. Hence parameter $\epsilon$ determines how fast the regret concentrates, and larger its value more 'risk-averse' is the algorithm. This motivates us to consider a variant of SEE that is more risk averse but at the cost of increased expected regret.

\subsection{Risk Averse Variant}
Our second algorithm-which we refer to as SEE2- is essentially same as SEE, except for the length of the exploration intervals which is exponential instead of super-exponential and does not depend on $\epsilon$. Specifically, we play the greedy arm $2^c$ times in cycle $c$. Compared to SEE, SEE2 spends significantly more time in the exploration intervals, and hence the probability that it makes error in the exploitation intervals is also significantly smaller and thus its regret concentrates around the expected regret faster.

\begin{theorem}
	\label{thm:ALGO2}
	Let the noise be $R$-sub-Gaussian and $\vth \in [-1, 1]^N$. Then, the expected regret of SEE2 is bounded as follows:
	\begin{equation}
	\label{eqn:RegAlg1}
	R_T(\text{SEE2} ) \leq 2R_m N\log^2T+ 4NR_m\gamma_2
	\end{equation}
		where  $\gamma_2$ is a constant that depends on noise parameter $R$ and the sub-optimality gap $\Delta$.
\end{theorem}

\section{Optimal Algorithm.}
We next obtain an optimal algorithm that achieves the lower bound in (\ref{thm:LowerBound}) within a constant factor when the sub-Gaussian parameter $R$ is known. 

\begin{minipage}{7.6cm}
	{\bf Algorithm-PolyLin:}\\
	In our next policy- which we refer as  \textbf{Poly}hedral-\textbf{Lin}ear-bandits we again split the time horizon into cycles consisting of an exploration interval followed by an exploitation interval as in SEE. As earlier, we index the cycles by $i$ and denote the exploration and exploitation intervals in cycle $i$ as $E_i$ and $R_i$, respectively. In the exploration interval $E_i$, we play each arm in $\mathcal{B}$ once. After $c$-cycles, using the rewards observed for each arm in $\mathcal{B}$ in the past $\{E_i, i=12,\cdots,c\}$ exploration intervals we compute ordinary least square (OLS) to estimate each component $\theta_n, n=1,2,\cdots,N$ separately, and obtain the estimate $\hat{\vth}(c)$ as follows.
	\beq
	\label{eqn:EstimateTheta2}
	\gthh_n(c)=\frac{1} {c} \sum_{i=1}^{c}  r_{t_{i,n}}/z_{n},
	\eeq
	Using $\hat{\vth}(c)$ as a proxy for $\vth$ we compute a greedy arm $\vx(c)$ and the sub-optimality gap $\hat{\Delta}(c)$ as follows.
	\[\hat{\Delta}(c)=\vx^\prime(c)\vth(c)-\max_{\vx \in \mathcal{C}\backslash \vx(c)}\vx^\prime\vth(c).\]
	In the exploitation interval $R_c$, we play $\vx(c)$ repeatedly for $2^{\kappa(c)c}$ times where $\kappa(c)$ is set to $a\hat{\Delta}(c)/2$, where $a=\min_{n}\OL{z}_n/R^2$. We repeat the process for each cycle. A formal description of PolyLin is given in the adjacent figure. 
\end{minipage}\hspace{.3 cm}
\begin{minipage}{5.8cm}
	\begin{algorithm}[H]
		\caption{PolyLin}
		\begin{algorithmic}[1]
			\STATE \textbf{Input:}
			\STATE $\mathcal{C}$: The polyhedron
			\STATE $R$: Noise parameter
			\STATE {\bf Initialization}
			\STATE  Compute the set $\mathcal{B}$
			\STATE $a:= \min_n \overline{z}_n^2 /R^2$
			\FOR {$i =1,2,\cdots$}
			\STATE \textbf{Exploration:}
			\FOR {$n=1\to N$}
			\STATE Play arm $z_n\ve_n \in \mathcal{B}$ \\ observe reward $r_{t_{i,n}}$
			\STATE $c=i$, Compute $\hat{\theta}_n(c)$ as in (\ref{eqn:EstimateTheta2})
			\ENDFOR 		
			\STATE $\hat{\vth}(c)\leftarrow (\hat{\theta}_1(c),\hat{\theta}_2(c)\cdots, \hat{\theta}_N(c))$
			\STATE $\vx(c)\leftarrow \displaystyle \arg\max_{\vx \in \mathcal{C}} \vx^\prime \hat{\vth}(c) $
			\STATE $\kappa(c)\leftarrow a\hat{\Delta}(c)/2$
			\STATE \textbf{Exploitation:}
			\FOR {$j=1 \to \lfloor 2^{\kappa(c) c} \rfloor$}
			\STATE Play arm \vx(c), observe reward
			\ENDFOR
			\ENDFOR
		\end{algorithmic}
	\end{algorithm}
\end{minipage}

\vspace{.5cm}
\noindent
Note that the exploration intervals of PolyLin are fixed length, whereas in SEE they are increasing as the the time progresses. Also, exploitation intervals in PolyLin are adaptive, whereas it is non-adaptive in SEE.

\begin{theorem}
	\label{thm:ALGO3}
	Let the noise be $R$-sub-Gaussian and without loss of generality assume $\vth \in [-1, 1]^N$. Then, the expected regret of PloyLin is bounded as follows:
	\begin{equation}
	\label{eqn:RegAlg3}
	R_T(PolyLin) \leq 2R_m N\frac{\log T}{\kappa}+ 4R_mN\gamma_3,
	\end{equation}
	where $R_m$ denotes the maximum reward.  $\gamma_3$ and $\kappa$ are constants that depends on noise parameter $R$ and the sub-optimality gap $\Delta$.
\end{theorem}

\section{Regret Analysis}
\label{sec:Analysis}
In this section we prove Theorem \ref{thm:ALGO1}, the proof of Theorem \ref{thm:ALGO2} follows similarly and omitted.  We first derive the probability of error in estimating each component of $\vth$ in each cycle. Note that in the exploration stage of each cycle $c$ we sample each arm $z_n\ve_n \in \mathcal{B}, i=1,2,\cdots,N$, 2 times more than that in the exploration stage of the previous cycle. Thus, we have $c^2$ plays of each arm $z_n\ve_n \in \mathcal{B}$ at the end of cycle $c$. The estimation error of component $\theta_n$ after $c$-cycles is given as follows:
\begin{lemma}
	\label{lma:EstimationErrorBound}
Let the noise be $R$-sub-Gaussian and $\delta>0$. In any cycle $c$ of both SEE and SEE2, for all $n=1,2,\cdots, N$ we have 	
	\begin{equation}
	\label{eqn:EstimatioErrorProb}
	P\left(\left|\gthh_n(c)-\gth_n\right|>\gd\right)\leq   2\exp\{ -c^2 \gd^2 \overline{z}_n^2 /2R^2 \}.
	\end{equation}
\end{lemma}
Note that larger the value of $\overline{z}_n$, the smaller the probability of estimation error is. The next lemma gives the probability that we play a suboptimal arm in the exploitation intervals of a cycle.

\begin{lemma}
	\label{lma:InfinityErrorBound}
For every cycle $c$, we have
\begin{itemize}
	\item[a.] Let $a:= \min_n \overline{z}_n^2 /R^2$. The estimation error is bounded as
	\begin{equation}
	\label{eqn:SupEstimationErrorProbability}
	\Pr\{{\|\vthh(c)-\vth\|_{\infty} > \eta}\}\leq 2N\exp\{-c^2\eta^2a\}. ,
	\end{equation}
	\item[b.] Let $h=\sup_{\vx \in \mathcal{C}} \|\vx\|_1$. The error in reward estimation is bounded as
	\beq 
	\label{eqn:LossInReward}
	\Pr\left(\exists \;\; \vx \in \mathcal{C} \;\;\text{such that} \;\; \left|\vthh^\prime(c) \vx-\vth^\prime \vx\right|>\eta \right) \le 2Ne^{-\frac{c^2 \eta^2a}{h^2}}.
	\eeq
	\item[c.] Probability that we play a sub-optimal arm is bounded as
	\beq
	\label{eqn:LinearityProperty}
	\Pr\left(\arg\max_{\vx\in \mathcal{C}} \vthh(c)^\prime \vx \neq\arg\max_{\vx\in \mathcal{C}} \vth^\prime \vx. \right) \le 2Ne^{-\frac{ac^2 \Delta^2/4}{h^2}}.
	\eeq
\end{itemize}

\end{lemma}

The proofs of Lemmas \ref{lma:EstimationErrorBound} and \ref{lma:InfinityErrorBound} are given in appendix. Recall that the number of extremal points is finite for the polyhedron $\mathcal{C}$ and $\Delta>0$. We use this fact to argue that whenever $\|\vthh(c)-\vth\|_{\infty} < \Delta/2$,  the greedy stage of the algorithm selects the optimal arm. This in an importation observation and follows from continuity property of optimal point in linear optimization theory \cite{Book_IntroductionLinear_BertsimasTsitsiklis}. Further, the probability of this event decays super-exponentially fast in our   policy implying that the probability that we incur a positive regret in the exploitation intervals is gets negligibly small over the cycles.
We compute the expected regret incurred in the exploration and exploitation intervals separately.
\subsection{Regret of SEE.} We analyze the regret in the Exploration and Exploitation phases separately as follows.\\
{\bf Exploration regret}: At the end of cycle $c$, each arm in $\mathcal{B}$ is played $\sum_{i=1}^c (2 i+1)=c^2$ times. The total expected regret from the exploration intervals after $c$ cycles is at most $Nc^2R_{m}$.\\
{\bf Exploitation regret}:  Total expected regret from the exploration intervals after $c$ cycle is
\begin{equation}
4NR_m \sum_{i=1}^c 2^{i^{2/(1+\epsilon)}}2^{-i^2a\Delta^2}=4NR_m \sum_{i=1}^c 2^{i^{2/(1+\epsilon)}-i^2a\Delta^2}\leq 4NR_m \gamma_2
\end{equation}
where $\gamma_2:=\sum_{i=1}^\infty 2^{i(i^{(1-\epsilon)/(1+\epsilon)}-c_1i \Delta^2/4)} $ is a convergent series.
\noindent
After $c$ cycles, the total number of plays is $T=\sum_{i=1}^{c} e^{i^{\frac{2}{1+\epsilon}}}+Nc^2 \geq e^{c^{\frac{2}{1+\epsilon}}}$ and we get $c^2 \leq \log^{1+\epsilon}T $. Finally,  expected regret form $T$-rounds is bounded as
\[R_T(SEE)\leq 2R_mN \log^{1+\epsilon} T + 4NR_m \gamma_2 =\mathcal{O}(N\log^{1+\epsilon} T).\]

\subsection{Regret of PolyLin.} We analyze the regret in the Exploration and Exploitation phases separately as follows.\\
{\bf Exploration regret}: After $c$ cycles, each arm in $\mathcal{B}$ is played $c$ times. The total expected regret from the exploration intervals after $c$ cycles is at most $NcR_{m}$.\\
{\bf Exploitation regret}:  Total expected regret from the explorations interval after $c$ cycles is
\begin{equation}
4NR_m \sum_{i=1}^c 2^{\kappa(i)i}2^{-ia\Delta^2}=4NR_m \sum_{i=1}^c 2^{i\kappa(i)-ia\Delta^2}\leq 4NR_m \sum_{i=1}^\infty 2^{ia\{ \hat{\Delta}^2(i)/2-\Delta^2\}}.
\end{equation}
Now consider the series $\gamma_3:=\sum_{i=1}^\infty 2^{ia\{\hat{\Delta}^2(i)/2-\Delta^2\}} $.
\begin{itemize}
	\item From Lemma \ref{lma:InfinityErrorBound}(a), $\vth(c) \rightarrow \vth$ as $c \rightarrow \infty $almost surely, we get $\hat{\vx}(c) \rightarrow \vx^*$ almost surely and  which in turn implies $\hat{\Delta}(c)\rightarrow \Delta$ almost surely.
	\item Then, for $0<\epsilon< \Delta^2/4$, the difference $ \hat{\Delta}(c)^2/2-\Delta^2\leq -\Delta^2/2 + \epsilon < 0$ for all but finitely many $c$. Hence, $\gamma_3$ is finite.  
\end{itemize}
After $c$ cycles the total number of plays is $T=\sum_{i=1}^{c} 2^{i\kappa(i)}+Nc \geq 2^{c\kappa (c)}$, and we get $c \leq \frac{\log T}{\kappa (c)}$. Finally,  expected regret form $T$-rounds, as $T \rightarrow \infty$, is bounded as
\[R_T(PolyLin)\leq 2R_mN \frac{\log T }{\kappa(c)}+ 4NR_m \gamma_3 .\]
Note that $ \hat{\Delta}(c)^2/2-\Delta^2 \geq -\Delta^2/2 -\epsilon $ for all but finitely many $c$. Then for sufficiently large $c$ we get $k(c)/a \geq \Delta^2/2-\epsilon \geq \Delta^2/4.$ Substituting in the last inequality we get
\[R_T(PolyLin)\leq 8R_mN \frac{\log T }{a\Delta}+ 4NR_m \gamma_3 =\mathcal{O}(N\log T).\]

\section{General Polyhedron}
\label{sec:GeneralPoly}
In this section we extend the analysis of the previous section to the case where origin is not an interior point of $\mathcal{C}$.

Analogous to set $\mathcal{B}$, we first define a set of arms that lie on the boundary of the polyhedron and these points are computed with respected to an interior point $\overline{\vx}$ of $C$ that we use as a proxy for origin. We use OPT-1 to find an interior point, whose smallest distance to boundaries along all the directions  $\{\ve_1, \ve_2,\cdots \ve_N \}$ is the largest.
The motivation to maximize the minimal distances to the boundaries comes from lemma \ref{lma:InfinityErrorBound}, where larger value of $a$ imply smaller probability of estimation error.

\begin{minipage}{7.5cm}
	{\bf OPT-1:}
\begin{equation*}
\begin{aligned}
&(\overline{\vx},\overline{\vy})=\arg \max _{\vx}  \min_{i} y_i \\
& \text{subjected to:}\\
& \mA\vx \leq \vb\\
& y_i \geq 0 \;\; \forall i=1,2,\cdots,N\\
& \mA (\vx + y_i\ve_i) \leq \vb \;\; \forall i=1,2,\cdots,N\\
& \mA (\vx - y_i\ve_i) \leq \vb \;\; \forall i=1,2,\cdots,N
\end{aligned}
\end{equation*}
\end{minipage}
\begin{minipage}{6cm}
	{\bf OPT-2:}
\begin{equation*}
\begin{aligned}
&(\overline{\vx},\overline{\vy})=\arg \max _{\vx, \vy,\alpha}  \alpha\\
& \text{subjected to:}\\
&  \alpha>0; \quad \mA\vx \leq \vb\\
& y_i -\alpha > 0 \;\; \forall i=1,2,\cdots,N\\
& \mA (\vx + y_i\ve_i) \leq \vb \;\; \forall i=1,2,\cdots,N\\
& \mA (\vx - y_i\ve_i) \leq \vb \;\; \forall i=1,2,\cdots,N
\end{aligned}
\end{equation*}
\end{minipage}

OPT-1 can be translated into an  equivalent linear progamme given in OPT-2 and hence the point $\overline{\vx}$ can be efficiently computed. We note that the set of points $\{\overline{\vx}+y_n\ve_n:n=1,2,\cdots,N\}$ need not all necessarily lie on the boundary. To see this, let the point $\overline{\vx}$ returned by OPT-1 is such that it is closer to the boundary along $i$th direction. Then the vector with all its component equal to  $y_i$ is a solution of OPT-1. To overcome this, we further stretch each point $\overline{\vx}+y_n\ve_n$ along the direction $\ve_n$ such that it hits the boundary. Let $\overline{z}_n=\arg\max_z \{|z| : z\ve_n \in C \}$. Finally, we fix the set of arms we use for explorations as $\overline{\mathcal{B}}=\{\overline{z}_n\ve_n + \overline{\vx}: n=1,2,\cdots,N \}$.

We are now ready to present an algorithm for linear bandits over for any polyhedra. For the general polyhedron, we use the SEE with the exploration strategy modified as follows.  In cycle $c$, we first play the arm $\overline{\vx}$ for $2c+1$ and then play each arm in $\overline{\mathcal{B}}$ $2c+1$ times as earlier. To estimate the component $\theta_n$, we average the difference in rewards observed from arms $\overline{\vx}+ \overline{z}_n\ve_n$ and  $\overline{\vx}$ so far.
From a straightforward modification of regret analysis of SEE, we can show that the expected regret of modified algorithm is upper bounded as $\mathcal{O}(N\log^{1+\epsilon} T)$ for all $\epsilon>0$.

The new algorithm required that we play the arm $\overline{\vx}$ along with the arms in $\overline{\mathcal{B}}$ in the exploration intervals to obtain estimate of $\vth$, and it increases the length of exploration intervals. However, it is possible that one can obtain estimates only by playing arms in $\overline{\mathcal{B}}$ provided we suitably modify the estimation method. More details are given in the appendix.
\section{Probability 1 Regret Bounds}
\label{sec:HighConfidence}
Recall the definiton of expected regret and regret in (\ref{eqn:pseudo}) and (\ref{eqn:regret}). In this section we show that with probability 1, the regret of our algorithms are within a constant factor from the their expected regret.

\begin{theorem}
With probability $1$,  $\overline{R}_T(SEE)$  is  $ \mathcal{O}(N\log^{1+\epsilon}T)$ and  $\overline{R}_T(SEE2)$  is  $ \mathcal{O}(N\log^2T) $.
\end{theorem}
{\bf Proof:} Let $\mathbb{C}_n$ denote an event that we select sub-optimal arm in the $n$th cycle. From Lemma \ref{lma:InfinityErrorBound}, this event is bounded as $\Pr\{\mathbb{C}_n\} \leq N \exp\{-\mathcal{O}(n^2)\}$. Hence $\sum_{n=1}^{\infty} \Pr\{\mathbb{C}_n\} < \infty$. Now, from application of Borel-Cantelli lemma, we get
$\Pr\{ \limsup_{n \rightarrow \infty}\mathbb{C}_n\}=0$, which implies that almost surely SEE and SEE2 play optimal arm in all but finitely many cycles. Hence the exploitation intervals contribute only a bounded regret. Since the regret due to exploration intervals is deterministic, the regret of SEE and SEE2 are within a constant factor from their expected regret with probability 1, i.e.,
$\Pr\{\exists \; C_1 \text{\;such that\;} \overline{R}_T(SEE)\leq R_T(SEE)+C_1\}$ and
$\Pr\{\exists \; C_2 \text{\;such that\;} \overline{R}_T(SEE2)\leq R_T(SEE2)+C_2\}$.
This completes the claim.

We note that the regret bounds proved in \cite{COLT08_StochasticLinearOptimization_DaniHayesKakad} hold with high confidence, where as ours hold with probability $1$ and hence provides a stronger performance guarantee.

\begin{figure}[!t]
	\vspace{-.5cm}
	\begin{minipage}{6.5cm}
		\centering
		\includegraphics[width=6.5cm, height=3.7cm]{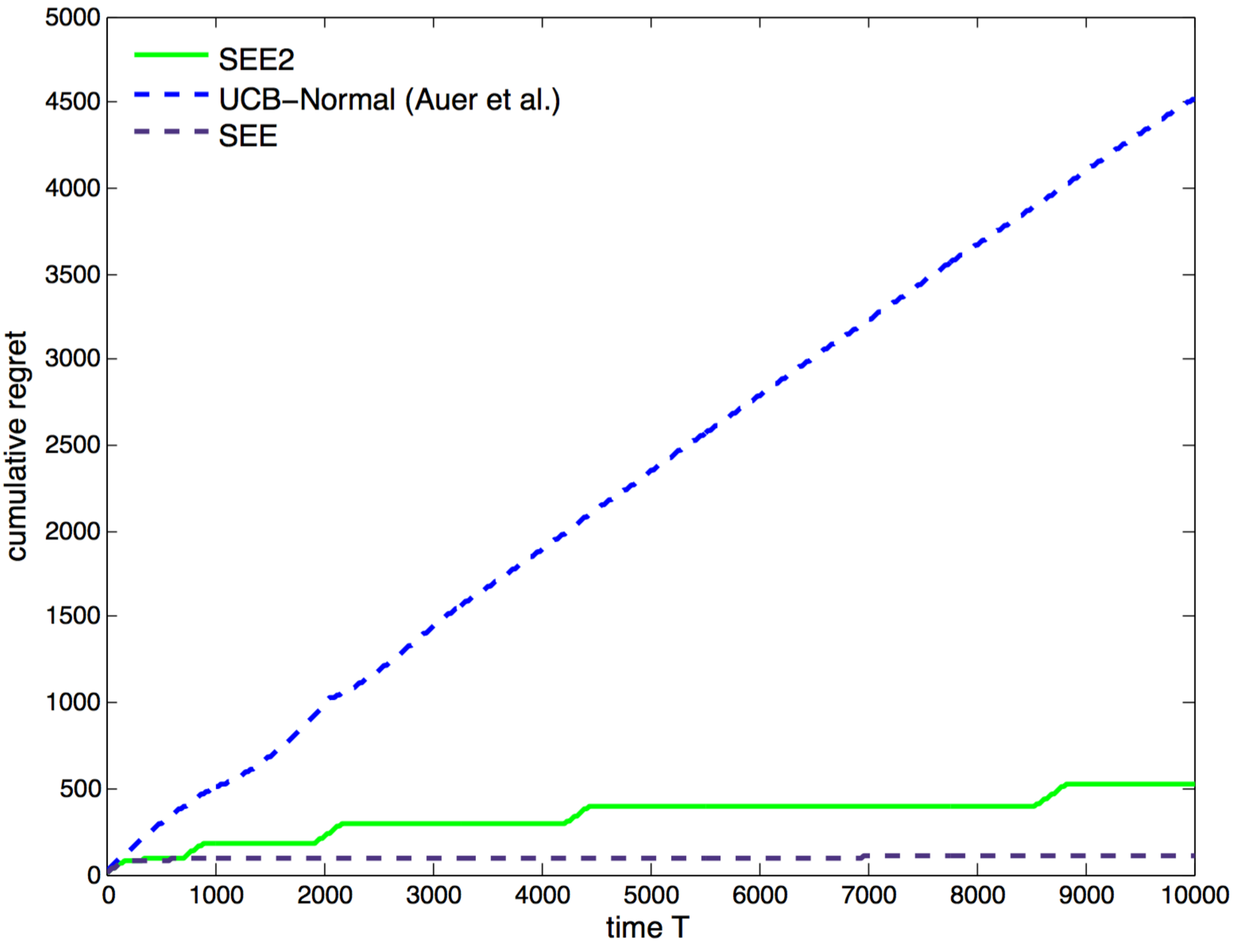}
		\label{fig:MultiArmed}
		\captionof{figure}{\small Regret comparison against multi-armed bandits, arms are corners of $10$-dim. hypercube.}
	\end{minipage}\hspace{1cm}
	\begin{minipage}{6cm}
		\centering
		\includegraphics[width=6.5cm, height=3.7cm]{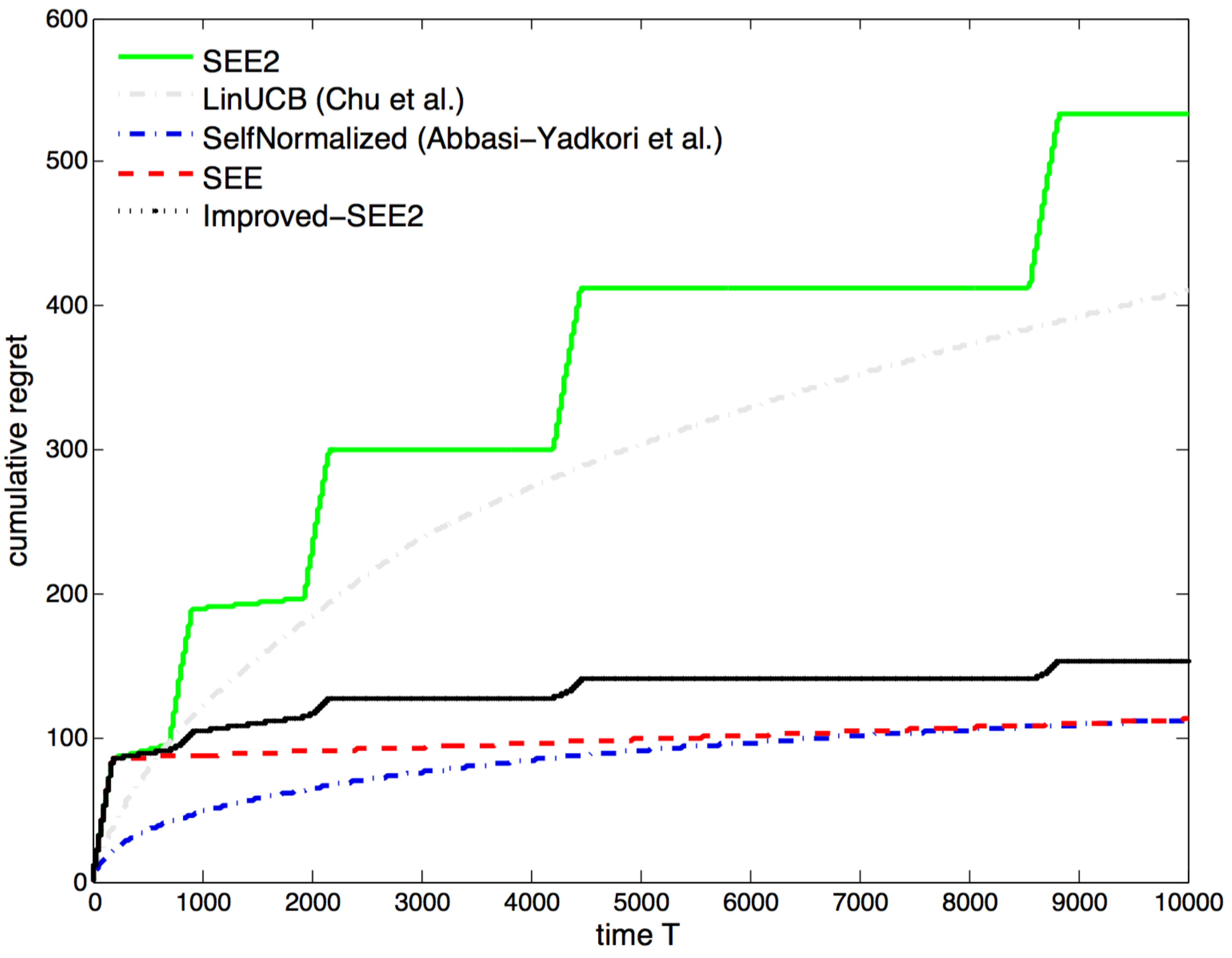}
		\label{fig:LinearBandit}
		\captionof{figure}{\small Regret comparison against linear bandit algorithms on $10$-dim. hypercube.}
	\end{minipage}
	\vspace{-.5cm}
\end{figure}

\vspace{-.3cm}
\section{Experiments}
\label{sec:Experiments}
In this section we investigate numerical performance of our algorithms against the known algorithms. We run the algorithms on a hypercube with dimension $N=10$. We generated $\vth\in [0,1]^N$ randomly and noise is zero mean Gaussian random variable with variance $1$ in each round. The experiments are averaged over $10$ runs. In Fig. 1 we compare SEE $(\epsilon=0.3)$ and SEE2 against UCB-Normal \cite{ML2002_FiniteTimeAnalysis_AuerCesaFischer}, where we treated each extremal point as an arm of an $2^N$-armed bandit problem. As expected, our algorithms perform much better. UCB-Normal need to sample each of the $2^N$ atleast once before it could start learning the right arm. Whereas, our algorithm starts playing the right arm after a few cycles of exploration intervals. In Fig. 2, we compare our algorithms against the linear bandits algorithm LinUCB and self-normalization based algorithm in \cite{NIPS2011_ImprovedAlgorithms_AbbasiPalSzepes}, which is labeled SelfNormalized in the figure. For these we set confidence parameter to $0.001$. We see that SEE beats LinUCB by a huge margin, but its performance comes close to that of SelfNormalized algorithm. Note that SelfNormalzed algorithm requires knowledge of sub-Gaussianity parameter $R$ of noises super. Whereas, our algorithms are agnostic to this parameter. Though, SEE2 seems to play the right arm in exploitation intervals, its regret performance is poor. This is due to increased number of exploration intervals, where no adaptation happens and a positive regret is always incurred.


The numerical performance of SEE2 can be improved by adaptively playing the arms in the exploration plays as follows, but at the increase cost of computations complexity. In each cycle $c+1$, we find a new set $\overline{\mathcal{B}}$ computed by setting $\overline{\vx}$ to $\vx(c)$, the greedy arm selected in the previous cycle, and play the new set arms as in the explorations intervals of the algorithm given for the general polyhedron. However, since $\vx(c)$ is an extremal points some of the $\overline{z}_n$'s are zero. To overcome this, we slightly shift the point $\vx(c)$ into the interior of the polyhedron along the direction $\vx(c)-\overline{\vx}$ and find a new set $\overline{\mathcal{B}}$ with respect to the new interior point. The regret of the algortihm based on this adaptive exploitation strategy is shown is Fig. 2 with label  'Improved-SEE2'. As shown, the modification improves performance of SEE2 significantly. In all the numerical plots, we initialized the algorithm to run from cycle number $5$.


\section{Conclusion}
We studied stochastic linear optimization over polyhedral set of arms with bandit feedback. We provided asymptotic lower bound for any policy and developed algorithms that are near asymptotically optimal. The regret of the algorithms grow (near) logarithmically in $T$ and its growth rate is linear in the dimension of the polyhedron. We showed that the regret upper bounds hold almost surely.
The regret growth rate of our algorithms is  $\log^{1+\epsilon}T$ for some $\epsilon>0$. It is interesting to develop strategies that work for $\epsilon=0$, while still maintain linear growth rate in $N$.
\bibliographystyle{IEEEtran}
\bibliography{bandits1,bandits2,bandits3}
\newpage
\appendix
\section*{Proof of Lemma \ref{lma:EstimationErrorBound}}
Let $\epsilon_{t_{i,n,j}}$ denote the noise in reward from playing $z_n\ve_n$ in phase $i$ for the $j$th time. We bound the estimation error as follows:
\begin{eqnarray}
\lefteqn{P\left(\left|\gthh_n(c)-\gth_n\right|>\gd\right)}\\
\label{eqn:Estimation1}
&=& P\left(\left|\sum_{i=1}^{c^2}\epsilon_{t_{i,n,j}}\right / c^2 z_n|>\gd\right) \\
\label{eqn:Multiply1}
&=&	P\left(s\left|\sum_{i=1}^{c^2}\epsilon_{t_{i,n,j}}\right |>s c^2 z_n \gd\right) \\
\label{eqn:Exponentiation1}
&=&	P\left(\exp \left \{s\left|\sum_{i=1}^{c^2}\epsilon_{t_{i,n,j}}\right |\right \}>\exp\{s c^2 z_n \gd\} \right) \\
\label{eqn:Symmetry1}
&\leq& 2P\left(\exp \left \{s \sum_{i=1}^{c^2}\epsilon_{t_{i,n,j}}\right \}>\exp\{s c^2 z_n \gd\} \right) \\
\label{eqn:MarkovInequality}
&\leq & 2\mathbb{E}\left[ \exp \left \{s \sum_{i=1}^{c^2}\epsilon_{t_{i,n,j}} \right \} \right]\exp\{-s c^2 z_n \gd\}\}\\
\label{eqn:ConditionalIndependence}
&\leq & 2\prod_{i=1}^{c^2}\mathbb{E}\left[ \exp \left \{s \epsilon_{t_{i,n,j}} \right \} | \mathcal{F}_{t-1}\right]\exp\{-sc^2 z_n \gd\}\}\\
\label{eqn:SubGaussianity1}
&\leq & 2\prod_{i=}^{c^2}\exp\{s^2\beta^2/2\}\exp\{-sc^2 z_n \gd\}\}\\
\label{eqn:FinalEstimate}
&= & 2\exp\{c^2(s^2\beta^2/2-s z_n \gd)\}\},
\end{eqnarray}
where (\ref{eqn:Estimation1}) follows from estimation step given in (\ref{eqn:EstimateTheta}). In (\ref{eqn:Multiply1}) and (\ref{eqn:Exponentiation1}) we exponentiated both sides within the probability functions after multiplying them by $s>0$.  (\ref{eqn:Symmetry1}) follows by applying union bound and using the symmetric property of the noise terms. In (\ref{eqn:MarkovInequality}) we applied the Markov inequality. In (\ref{eqn:ConditionalIndependence}) we aplied conditional independence property of the noise. (\ref{eqn:SubGaussianity1}) follows by applying the definition of sub-Gaussian property.

Note that upper bound in (\ref{eqn:FinalEstimate}) holds for all $s>0$ and is minimized at $s^*= \frac{\gd z_n}{\beta^2}>0$. Finally, the lemma by substituting $s^*$ in  (\ref{eqn:FinalEstimate}).

\section*{Proof of Lemma \ref{lma:InfinityErrorBound}}
\subsection*{Part a:}
We bound the estimation error as follows:
\begin{eqnarray}
\lefteqn{\Pr\left(\left \|\vthh(c)-\vth\right \|_{\infty} >\eta\right) } \\
&\le& \Pr\left(\exists n: \left |\gthh_n(c)-\gth_n\right | >\eta\right) \\
\label{eqn:UnionBound}
&\le& \sum\limits_{n=1}^{N}\Pr\left( \left |\gthh_n(c)-\gth_n\right | >\eta\right) \\
\label{eqn:EstimationError1}
&\le& 2Nc_1e^{-a c^2 \eta^2}.
\end{eqnarray}
In (\ref{eqn:UnionBound}) we applied the union bound result and in (\ref{eqn:EstimationError1}) we applied (\ref{eqn:EstimatioErrorProb}).

\subsection*{Part b:}
For all $\vx \in \mathcal{C}$, we have 
\begin{equation}
\label{eqn:RewardBound}
|\vx^\prime\vth(c)-\vx^\prime\vth|\leq \|\vth(c)-\vth\|_\infty\|\vx\|_1.
\end{equation}
Define events $\mathbb{A}=\{\exists \;\vx\; \text{such that} |\vx^\prime\vth(c)-\vx^\prime\vth| > \eta \}$ and $\mathbb{B}=\{ \|\vth(c)-\vth\|_\infty h > \eta \}$. The last inequality implies $\Pr\{\mathcal{A}\} \leq \Pr\{\mathcal{B}\}$. The claim follows from part-a of the lemma.
\subsection*{Part c:}
Suppose $\vy \neq \vx^*$, where $\vx^*$ is the optimal arm, such that $\vth^\prime(c) \vy > \vth^\prime(c) \vx^*$. Then, since $\vth^\prime \vx^*- \vth^\prime \vy \geq \Delta$ we must have that either $|\vth^\prime \vx^* - \vth^\prime(c)\vx^*| \geq \Delta/2$ or $|\vth^\prime(c)\vy- \vth^\prime\vy|\geq \Delta/2$, otherwise we cannot close the gap. Hence, if the greedy selection in cycle $c$ is not $\vx^*$, it implies that there exists a $\vx \in \mathcal{C}$ such that $|\vth^\prime(c)\vx-\vth\vx|>\Delta/2$. From part-b this probability is bounded as $2N\exp\{-ac^2\eta^2/h\}$, where $\eta=\Delta/2$. This completes the proof.
\section*{Estimation in the case general polyhedron}
Let $\OL{\vx}_i=\OL{\vx} + \A_i\ve_i$. Let $\hat{r}_i(c):= \frac{1}{c^2} \sum_{i=1}^{c}\sum_{j=1}^{2c+1}r_{t_{i,n,j}}$ denote the average of the reward obtained from arm $\OL{\vx}_i$ till end of phase $m$. At the end of phase $m$, we estimate $\BS{\theta}$ as follows:

\[\hat{\BS{\theta}}(c)=\left (\BS{1}\OL{\vx}^\prime + \BS{D}(\A)\right )^{-1}\hat{\BS{r}}(c),\]
where $\BS{\A}$ denote the diagonal matrix with diagonal elements as $\A$ and $\hat{\BS{r}}(c)$ is the vector with $i$th component as $\hat{r}_i(m)$. By applying matrix inversion lemma we get

\[\hat{\BS{\theta}}(c)= \left(\BS{D}^{-1}(\A) -\frac{\BS{D}^{-1}(\A) \BS{1}\OL{\vx}^\prime\BS{D}^{-1}(\A) }{\OL{\vx}^\prime\BS{D}^{-1}(\A) \BS{1}}\right)\]
After simplification, for each $i=1,2,,\cdots,N$ we have

\[\hat{\theta}_i(c)=\frac{1}{\alpha_i}\left(\hat{r}_i(c)-\frac{\sum_{j=1}^N (\OL{x}_j/\alpha_j)\hat{r}_j(c)}{\sum_{l=1}^N \OL{x}_l/\alpha_l}\right)\]
Substituting the reward from arm $\OL{\vx}_i$, i.e.,
\[r_{\OL{\vx}_i}=\OL{\vx}^\prime\BS{\theta}+ \alpha_i \theta_i + \epsilon\] and further simplifying we get

\[\hat{\theta}_i(c)=\frac{1}{\alpha_i}\left(\alpha_i\theta_i- \OL{\vx}^\prime \BS{\theta} + \sum_{j=1}^{N} \beta_j \hat{\epsilon}_j(c)\right)\]
where $\beta_j=\frac{\OL{x}_j}{\alpha_j}$ and $\hat{\epsilon}_j(m)$ is the noise average from playing arm $\OL{\vx}_i$.
\end{document}